\documentclass[10pt,twocolumn,letterpaper]{article}
\usepackage[letterpaper,margin=0.75in]{geometry}
\usepackage[T1]{fontenc}
\usepackage{newtxtext}
\usepackage[hyphens]{url}
\usepackage{graphicx}
\usepackage{natbib}
\usepackage{caption}
\usepackage{algorithm}
\usepackage{algorithmic}
\usepackage{booktabs}
\usepackage{amsmath}
\usepackage{amssymb}
\newcommand{\bvec}{\mathbf{B}}
\newcommand{\method}{BAP-SQL}

\title{BAP-SQL: Budget-Aware Observation Planning for Agentic Text-to-SQL}
\author{
	{\fontsize{12}{15}\selectfont\bfseries
	Chong Peng\textsuperscript{1},
	Pin Qian\textsuperscript{2},
	Su Wang\textsuperscript{2},
	Yihang Chen\textsuperscript{3},
	Varun Sah\textsuperscript{1}}\\[0.4em]
	{\fontsize{9}{11}\selectfont\normalfont
	\textsuperscript{1}Microsoft \quad
	\textsuperscript{2}Carnegie Mellon University\\
	\textsuperscript{3}Georgia Institute of Technology}
}
\date{}

\makeatletter
\renewcommand{\@maketitle}{%
	\newpage
	\null
	\vskip 2em%
	\begin{center}%
		{\fontsize{14}{16}\selectfont\bfseries \@title\par}%
		\vskip 0.75em%
		{\@author\par}%
	\end{center}%
	\par
	\vskip 1.5em%
}
\makeatother

\begin{document}
\maketitle

\begin{abstract}
Tool-using agents do not merely consume observations: their actions determine
what arrives next. In agentic text-to-SQL, a broad query can spend context and
database work before useful evidence appears, while post-hoc compression cannot
recover omitted rows or expended work. We present \method{}, which treats
observation formation as a budget-control stage: it estimates query risk,
rewrites SQL when useful, and delegates hard limits to an independent runtime
shield. Across general 4B, specialized FINER-SQL 4B, and 7B backbones,
\method{} improves tight-budget success. On the primary BIRD-derived setting,
it gains 3.4/3.6 percentage points over matched SFT while using 4.5/5.0\%
fewer tokens. Matched retraining and task-level transfer associate the gain
with policy-visible planning and budget-sensitive rescue. The benefit
attenuates as model capability and budget increase, reverses at the loosest
setting, and does not reduce database work.

\end{abstract}

\section{Introduction}

Tool-using agents do not simply process observations; they choose actions that
create them. Consider an analytics agent asked which product category generated the most
revenue. A broad join returns thousands of rows, the interface truncates the
result, and the agent continues from an incomplete observation. Summarization
can reduce the text shown to the model, but it cannot undo the query or recover
evidence that never entered the context. In agentic SQL, query choice controls
result cardinality, observation size, and the information available for later
reasoning.

Recent text-to-SQL systems use execution feedback and reinforcement learning to
improve multi-turn correction and search
\cite{sqlr12025,rewardsql2025,mtirsql2025,sqltrail2026}. A parallel line of
work controls tool calls, reasoning length, or context growth
\cite{otcpo2025,alp2025,l1lcpo2025,tale2025,contextbudget2026}. Two gaps
remain. First, post-hoc context control cannot refund database work or recover
rows omitted by the original query. Second, correctness-oriented SQL training
does not decide whether the next observation is affordable. Our key insight is
that the SQL action itself is a budget-control decision: a selective aggregate
changes result cardinality, result tokens, and downstream evidence before the
observation exists.

\begin{figure*}[t]
\centering
\includegraphics[width=0.96\textwidth]{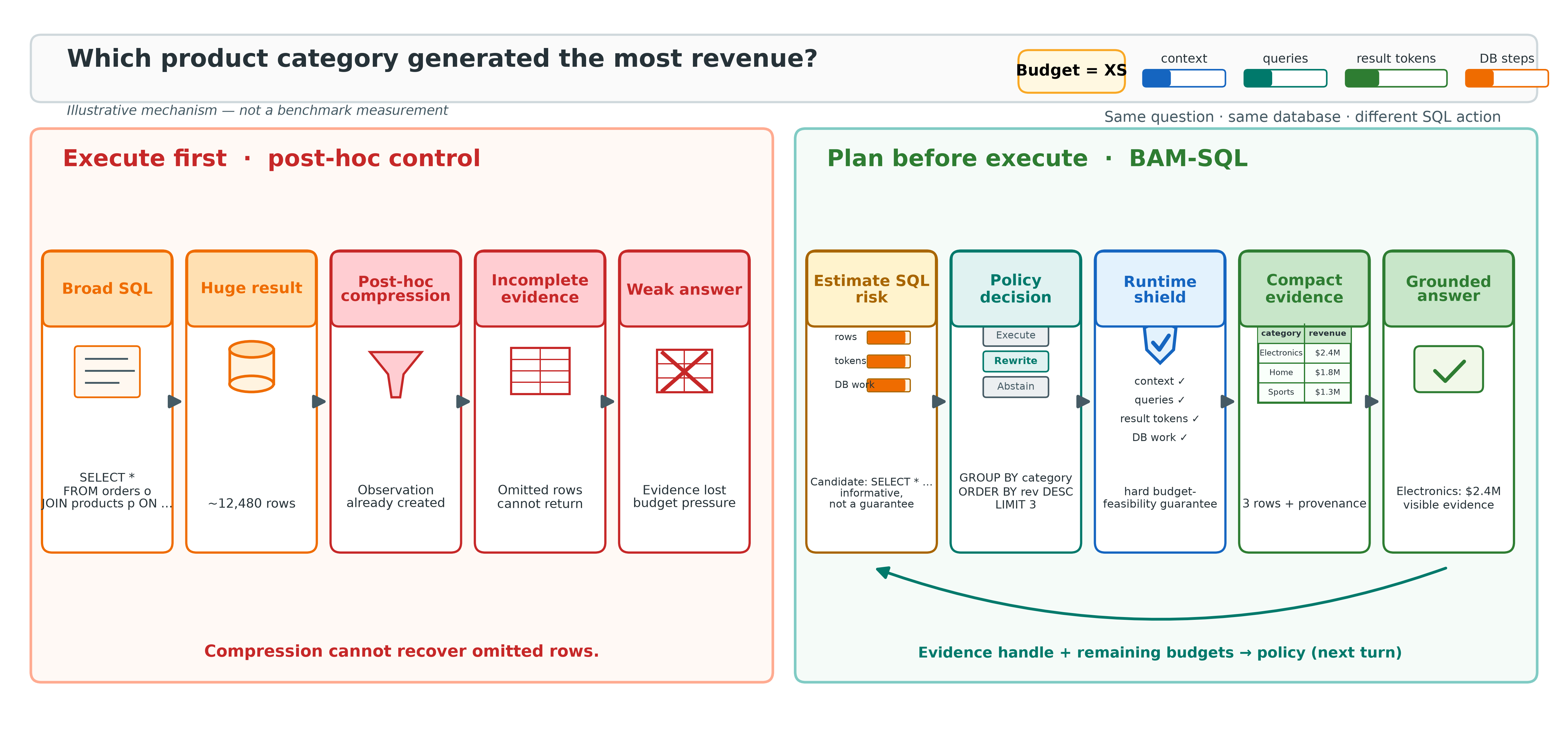}
\caption{Post-hoc control acts after a broad SQL query has created an
oversized observation. \method{} instead estimates the candidate, chooses
whether to rewrite it, executes under a runtime shield, and returns compact
visible evidence. The schematic is not a benchmark result; all quantitative
measurements follow the Experimental Setup protocol.}
\label{fig:overview}
\end{figure*}

\method{} operationalizes this insight. At every turn, the policy sees the
remaining budget and may inspect the schema, estimate a proposed query,
rewrite it, execute it, manage evidence, answer, or abstain. Cost estimates
guide the policy, while an independent runtime shield enforces hard limits.
The same policy is trained across four budget levels and adapts its actions to
the remaining resources. Figure~\ref{fig:overview} contrasts this
plan-before-execution loop with post-hoc control.

A deterministic rewriter is not sufficient for this setting. Whether an
estimate is worth requesting, whether a broad query should be replaced by an
aggregate, and whether more evidence is needed depend on the question,
previous observations, and remaining budget. Reinforcement learning provides
a natural way to learn these sequential choices from task outcomes while
assigning zero reward to wrong or infeasible trajectories.

This work makes four contributions. First, it identifies observation formation
as a distinct budget-control stage for tool-using agents and formulates its SQL
instance under joint context, query, result-token, and database-work budgets.
Second, it introduces a closed-loop policy that estimates query risk, chooses
whether to issue a different evidence query, manages recoverable evidence, and
separates planning information from hard enforcement. Third, matched
retraining, interface-matched SQL-RL, and a cost-matched placebo distinguish
planning, reward shaping, and inference-time estimate use. Fourth, experiments
across general 4B, specialized FINER-SQL 4B, and 7B generators characterize a
consistent regime: gains are largest when observation budgets are tight and
attenuate as generator capability and available resources increase.

\section{Related Work}

\paragraph{SQL reinforcement learning and interactive text-to-SQL.}
BIRD provides a realistic execution-based benchmark for text-to-SQL
\cite{bird2023}. Recent systems optimize SQL generation with outcome rewards,
process rewards, or multi-turn execution feedback
\cite{execguided2018,sqlr12025,rewardsql2025,mtirsql2025,sqltrail2026,reexsql2025}. MARS-SQL
extends this direction to trainable multi-agent workflows
\cite{marssql2025}. BIRD-INTERACT further emphasizes dynamic interaction rather
than one-shot generation \cite{birdinteract2026}. These systems learn how to
improve a query through feedback. \method{} focuses on a complementary
question: how to change the query before execution when its observation would
consume the remaining budget.

\paragraph{Budget-aware agents and adaptive computation.}
Tool-use and reasoning systems increasingly expose finite resources to the
policy. Search-R1 trains long-horizon search behavior with reinforcement
learning \cite{searchr12025}. OTC-PO rewards efficient tool use only when the
task is solved \cite{otcpo2025}; ALP, L1, and TALE adapt reasoning to a token
budget or input difficulty \cite{alp2025,l1lcpo2025,tale2025}; BAGEN predicts
remaining-budget intervals and learns when to stop \cite{bagen2026}.
ContextBudget formulates context compression as a budget-conditioned
sequential decision problem and trains BACM-RL to decide when and how much
history to compress \cite{llmlingua2023,contextbudget2026}. Its central empirical claim is
higher task performance under both fixed and progressively tighter context
budgets. LLMLingua compresses selected prompt content and ContextBudget manages
acquired interaction history; \method{} instead changes the SQL action that
produces a future observation.

\paragraph{Database cost control and runtime safety.}
Query optimizers estimate cardinality and cost before execution
\cite{learnedcard2019,optimizerSurvey2021,lpbound2025}, while SQLGovernor uses DBMS feedback for
SQL correction and rewriting \cite{sqlgovernor2025}. Runtime shielding and
database admission control provide a separate enforcement layer
\cite{shielding2018}. Cost-aware routing and accounting have also begun to
appear in text-to-SQL evaluation: EllieSQL studies routing
\cite{elliesql2025}, while CostSQL studies realized cost
\cite{costsql2026}. \method{}
connects these ideas to an agent policy: the estimate is exposed as decision
information, whereas the runtime shield remains responsible for feasibility.
Evidence storage follows the broader line of virtual context and semantic
memory \cite{memgpt2023,agentsm2026}. Unlike a database optimizer, the policy
may choose a non-equivalent evidence query because the observation and
downstream reasoning process are jointly budgeted; learned optimizers instead
select lower-cost physical plans for equivalent SQL.

\section{Problem Formulation}

An episode contains a question $q$, schema $\mathcal{S}$, database $D$, and
joint budget
\[
  \bvec=[B_{\mathrm{ctx}},B_{\mathrm{query}},
  B_{\mathrm{result}},B_{\mathrm{db}}].
\]
The state includes the active transcript, remaining budget, prior estimates,
evidence blocks, and executed SQL. The policy emits one typed action at each
turn from
\[
\begin{aligned}
\mathcal{A}=\{&\textsc{Inspect},\textsc{Estimate},\textsc{Rewrite},
\textsc{Execute},\\
&\textsc{Manage},\textsc{Answer},\textsc{Abstain}\}.
\end{aligned}
\]
A SQL action is therefore not merely a retrieval call. It changes the next
observation through its result cardinality and projected structure.

\begin{table}[t]
\centering
\footnotesize
\begin{tabular}{@{}lrrrrr@{}}
\toprule
Budget & Context & Queries & Result tok. & VM steps & Turns \\
\midrule
XS & 2,400 & 1 & 80 & 250,000 & 5 \\
S  & 3,500 & 2 & 250 & 600,000 & 7 \\
M  & 6,000 & 4 & 800 & 1,500,000 & 10 \\
L  & 11,000 & 8 & 2,500 & 4,000,000 & 14 \\
\bottomrule
\end{tabular}
\caption{Frozen joint budget ladder: live context tokens, executed queries,
cumulative admitted result tokens, SQLite VM steps, and turn ceiling.}
\label{tab:budgets}
\end{table}

The channels have different accounting semantics. $B_{\mathrm{ctx}}$ limits
the live prompt at each turn, so archiving or compression restores context
headroom. Executed queries, admitted result tokens, and deterministic SQLite
VM steps are cumulative. Estimate and syntax-check actions consume prompt
tokens and measured database work but do not increment the query channel;
estimate, rewrite, and evidence-management actions consume a turn. A
preflight-rejected execution consumes a turn but no query because SQL never
runs. Catalog construction is shared offline preprocessing and is not charged
to individual episodes.

We use a small, stable metric vocabulary. \emph{Budgeted Success} means that
the final answer is correct and the episode respects all budgets.
\emph{Official BIRD EX} is the standard unbudgeted execution-accuracy anchor.
\emph{Evidence Support} measures whether cited, policy-visible rows justify the
answer. \emph{Total tokens} are cumulative prompt, completion, result, and
controller tokens across the episode; they are never subtracted.
\emph{Active context} is the live prompt buffer at a turn and can decrease
after compression or archiving.

The \textsc{Answer} action emits a typed scalar, list, or ordered list.
Correctness uses task-specific normalization and numeric tolerance; lists use
multiset equality, with order enforced only for ordered-list tasks. Malformed
answers are incorrect. Let $y_i$ be this answer and let $z_i=1$ when every
channel respects its own level or cumulative semantics. For task $i$,
\[
  s_i=\mathbb{1}[y_i=y_i^\star]\mathbb{1}[z_i=1].
\]
The shield makes hard breaches unreachable during normal operation; budget
pressure instead appears through rejected actions, capped observations, and
incorrect answers. We retain $z_i$ as an audit invariant for harness or
accounting failures.

\emph{Frontier Score} is the trapezoidal mean over the four equally spaced
ordinal budget levels, reported in percentage-point units rather than as a
probability.

\section{BAP-SQL}
\label{sec:method}

\method{} separates policy decisions from hard enforcement
(Figure~\ref{fig:overview}). At turn $t$, the policy observes the question,
schema summary, remaining budget $\bvec_t$, prior estimates, and stored-evidence
references. It emits one typed action---inspect, estimate, rewrite, execute,
manage evidence, answer, or abstain---and receives an updated ledger. The same
policy serves all budgets: tight regimes favor one compact aggregate, whereas
larger budgets permit drill-down and corroboration.

\paragraph{Pre-execution query planning.}
Given a proposed SQL query, \texttt{estimate\_query} combines
\texttt{EXPLAIN QUERY PLAN}, a zero-row probe, and per-database catalog
statistics. It returns p50 and p95 estimates for rows, result tokens, and
database work. Catalog statistics provide cardinalities, distinct counts, and
serialized widths; SQL-shape features mark joins, filters, aggregates, limits,
and projections. Calibration factors are fitted on disjoint databases. A
\texttt{LIMIT 0} probe validates bindings and output columns without admitting
rows. The policy may execute, rewrite with aggregation or selective predicates,
request another evidence query, or abstain. Rewrites need not preserve the
discarded candidate's semantics; correctness and visible support are judged
against the user question.

\paragraph{Runtime shielding.}
Before execution, the environment checks the remaining ledger and configures
VM-step, row, byte, and result-token caps. Estimates are used to expose risk
and reject clearly infeasible actions; execution-time caps provide the hard
guarantee. Accounting continues after visible output is capped, so truncation
does not refund the scan. Syntax checks and estimates consume context and
database work but not the result-producing query channel.

\paragraph{Recoverable evidence.}
Query results are stored as evidence blocks containing rows, SQL provenance,
profiles, truncation status, and retrieval references. \textsc{Archive}
removes exact rows from active context, \textsc{Compress} replaces them with a
deterministic profile, and \textsc{Discard} makes them unavailable. Support is
computed only from visible rows. Restoring or fetching rows consumes result
tokens again and returns them to active context; cumulative charges are never
refunded.

\paragraph{Training.}
Nine programmatic workflows provide action syntax and initial budget behavior,
including direct aggregation, estimate--rewrite, drill-down, evidence
compression, and justified abstention. We retain 8,192 turn-level SFT pairs and
train rank-32 LoRA adapters \cite{lora2021}. During RL, each task--budget prompt
is sampled eight times; groups with 2--6 successes provide the online
difficulty-filtered training signal \cite{odf2026}.

\paragraph{RL objective and configuration.}
We use correctness-gated efficiency shaping rather than an explicit
token, context, or database-work reward. For a correct and feasible trajectory with
evidence score $e$, executed queries $m$, and the minimum successful group
query count $n$,
\[
R=\mathbf{1}[\mathrm{correct}\wedge\mathrm{feasible}]\,(1+0.4e)\,r_{\mathrm{tool}}(m,n).
\]
Wrong answers and hard-budget violations receive zero, so efficiency only ranks
correct trajectories; a cheap wrong answer cannot receive a positive reward.
The deterministic evidence score lies in $[0,1]$ and combines valid citations,
provenance, visible answer support, and corroboration; no learned judge or
hidden rows are used.
The tool factor follows the correctness-gated multiplicative form used by
OTC-PO, with $n$ taken from the smallest successful query count in the same
rollout group \cite{otcpo2025}: for $n>0$,
$f(m,n)=2mn/(m+n)$ and
$r_{\mathrm{tool}}=\sin\!\left(f(m,n)\pi/(2n)\right)$.
For the zero-query fallback we use the corresponding OTC-PO cosine branch.
Budget state remains visible to the policy, and the runtime shield is
unchanged. The no-efficiency-shaping ablation keeps the
same state, evidence bonus, and shield but sets $r_{\mathrm{tool}}=1$.
Matched SQL-RL instead uses pure correctness reward, removing both the evidence
bonus and tool-efficiency factor. We optimize three adapters with
group-relative policy optimization \cite{deepseekmath2024,dapo2025};
configuration details appear in Section~\ref{sec:experiments}.

\section{Experimental Setup}
\label{sec:experiments}

\paragraph{Evaluation at a glance.}
The primary endpoint compares \method{} RL with matched SFT at XS/S using an
ordered quality-then-token test. Interface-matched SQL-RL is the closest
trainable control, adapted BACM-RL is the external post-hoc comparator, and
FINER-SQL tests transfer to a stronger SQL-specialized backbone.

\paragraph{Data.}
The BIRD-derived source contains 8,477 tasks from 69 databases
\cite{bird2023}: 7,496 train and 981 held-out validation tasks after
database-level splitting. Dev contains 1,523 executable tasks from 11 disjoint
databases, with no database or exact-question overlap. Budgeted Success uses
task-specific scalar/list answers; Official BIRD EX remains an external anchor
on all 1,534 official dev records.

\paragraph{Budgets and models.}
We evaluate four joint budgets (Table~\ref{tab:budgets}) using Qwen3.5-4B
with reasoning mode disabled \cite{qwen35model}. The ladder was chosen from training-source
profiling to span severe through moderate serving constraints and was frozen
before dev evaluation. Conditions share tasks, matched SFT initialization,
online filtering, decoding, 120 updates, three seeds, syntax checks, and the
runtime shield; controllers pay their prompt and output tokens. The ladder
varies all scientific channels together. Mixed single-channel vectors are
out-of-distribution sensitivity probes, and a common-turn control checks parity.

\paragraph{Baselines.}
Raw and Dashboard isolate prompting and budget visibility; Dashboard has no
estimator or rewrite action. Adaptive post-hoc control learns truncation,
summarization, profiling, or a fixed limit after execution. Interface-matched
SQL-RL shares the budget state, estimator, rewrite, evidence memory, and
training protocol but uses pure correctness reward. Context-only RL freezes the
matched-SFT SQL generator and learns evidence management with the full
\method{} reward. Adapted BACM-RL uses its released compression policy with the
same frozen SQL generator. Its prompts, summaries, and outputs are charged, but
the frozen generator makes this a secondary comparison; SQL-RL is the tighter
trainable-SQL control.

\paragraph{Metrics and inference.}
\emph{Budgeted Success} is the primary quality measure. We additionally report
Official BIRD EX, total tokens, active context, queries, result tokens, and
database work. Quality and cost are tested in order: the primary XS/S
comparison against SFT first uses a five-percentage-point non-inferiority
margin, then tests paired token reduction. Ten thousand
paired bootstrap replicates resample the 11 databases while retaining all
task--seed observations; intervals therefore condition on the three adapters.
Database-$t$, leave-one-database-out, per-seed, macro, and sign-test analyses
check the direction. Frontier Score is a descriptive trapezoidal mean over the
four ordinal budgets. The primary gate is conjunctive at $\alpha=0.05$; other
intervals are exploratory and unadjusted.

\paragraph{Strong-backbone transfer.}
We evaluate FINER-SQL-4B-BIRD \cite{finersql2026} under the same budgets.
Its frozen one-shot checkpoint reaches 41.56/48.33/51.54/55.88\% at XS/S/M/L
and 66.82\% Official EX. Comparing the same one-shot SQL across budgets
partitions tasks into always-correct, budget-sensitive, and persistent-under-L
groups. We train matched FINER-initialized SQL-RL, BACM-RL, and \method{}
adapters; frontier cells average three seeds, while Figure~\ref{fig:finer-rescue}
shows the median-seed transition matrix.

\paragraph{Reproducibility.}
SFT runs for one epoch ($10^{-4}$ learning rate, effective batch 128, maximum
length 8,192). RL uses learning rate $10^{-6}$, temperature 1.0, top-$p$
0.95, 16 groups of eight trajectories per update, 120 updates, and three
seeds. Primary training and failed launches consume approximately 190 A100
GPU-hours. We release prompts, configs, hashes, trajectories, and analysis.

\paragraph{Ablations.}
The locked retraining suite is independent of the main frontier. It separately
trains a full-interface SFT checkpoint and an action-matched no-planning SFT
checkpoint on the same task--budget order, 8,192 turn pairs, maximum length,
and update budget. No-planning retains \textsc{Inspect}, \textsc{Execute},
evidence management, \textsc{Answer}, and \textsc{Abstain}, but removes
policy-visible p50/p95 estimates and \textsc{Rewrite}. The runtime shield and
its preflight estimator remain unchanged; the policy observes only ordinary
rejection outcomes, not estimator warnings. Pair count and maximum sequence
length are matched, but exact token exposure is not forced equal. The two SFT
initializations have the same rounded aggregate performance, 31.4/34.3 at
XS/S, before their respective RL stages.

We retrain three no-planning adapters with the full reward and three with pure
correctness reward, using the frozen prompt pool, online filter, optimizer, and
120-update budget. Together with the corresponding planning-enabled runs, this
forms a $2\times2$ factorial over policy-visible planning and reward shaping.
All effects in the left panel of Table~\ref{tab:mechablate} are computed within
this locked suite using paired database bootstrap intervals.
Fixed-adapter no-rewrite/no-interface
interventions diagnose operational dependence in the separate main run. A
cost-matched placebo permutes only the policy-visible p50/p95 payload while the
shield retains true estimates; it tests inference-time reliance, not retrained
causality.

\section{Results}

\begin{figure*}[t]
\centering
\includegraphics[width=0.96\textwidth]{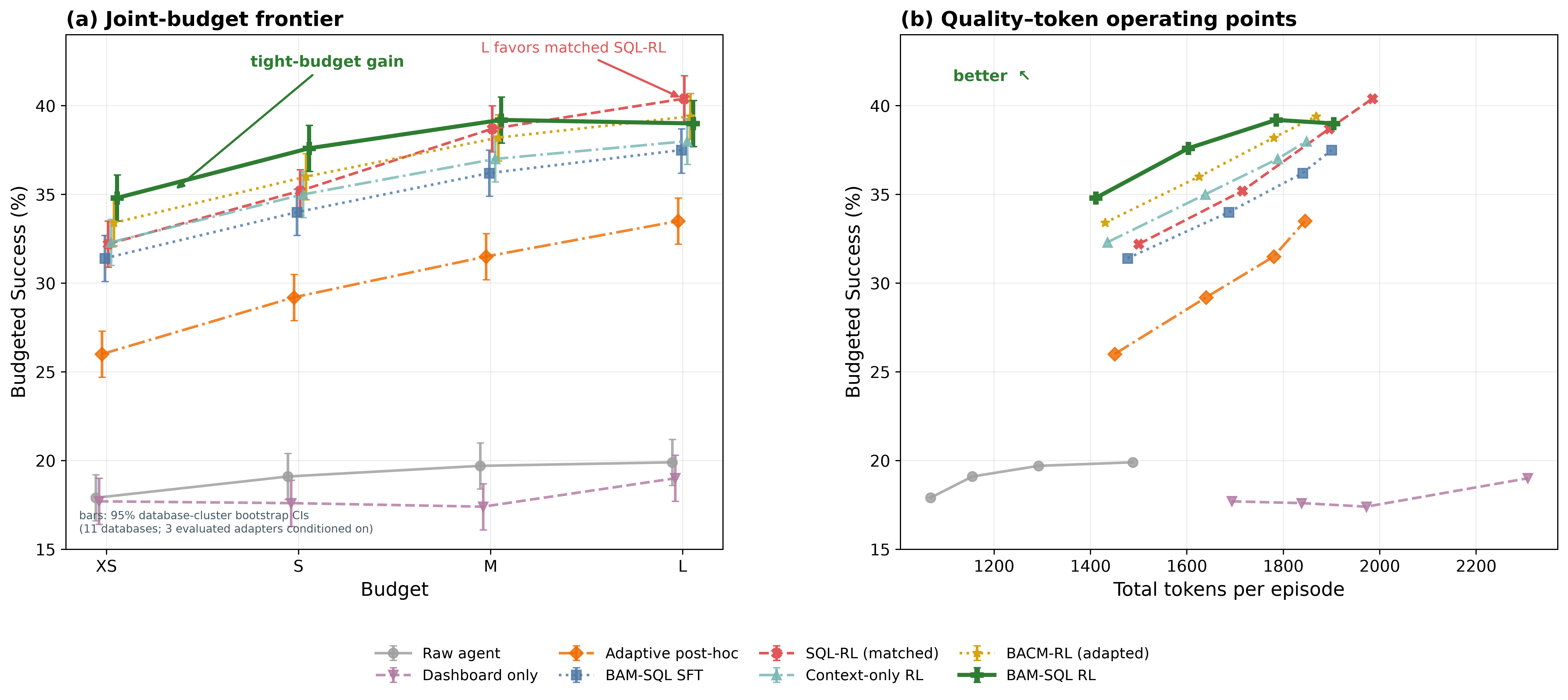}
\caption{Budget conditioning changes the tight-budget operating point. Left:
Budgeted Success over the four budgets. Right: the corresponding
quality--token trajectories. Markers pool three adapters; vertical bars are
95\% database-cluster bootstrap intervals over 11 databases, conditional on
those adapters. \method{} has the highest observed means at XS/S; secondary L
means favor matched SQL-RL.}
\label{fig:frontier}
\end{figure*}

\paragraph{RQ1: Tight-budget frontier.}
\method{} has the highest observed Budgeted Success at XS/S and the highest
Frontier Score (Table~\ref{tab:frontier}; Figure~\ref{fig:frontier}). Relative
to SFT, the XS/S gains are 3.4 and 3.6 percentage points. The three adapter
Frontier Scores are 38.6, 37.9, and 37.2 (sample standard deviation 0.70).
Relative to adapted BACM-RL, the gains are 1.4 and 1.6 percentage points, with paired
intervals $[0.8,2.0]$ and $[0.9,2.3]$. The advantage narrows as budgets loosen:
against interface-matched SQL-RL, the XS/S differences are $+2.6$
$[1.8,3.4]$ and $+2.4$ $[1.6,3.2]$.
At L, matched SQL-RL is 1.4 percentage points higher than \method{}
($[-2.5,-0.3]$ for \method{}--SQL-RL). Official BIRD EX is 42.0\% for
\method{} RL and 41.9\% for SFT; these are descriptive unbudgeted anchors.
Across seeds, the paired XS/S gains range from 2.6--4.1 and 2.8--4.5 points.
Database-macro gains are 3.4/3.6 points, and leave-one-database-out ranges are
$[3.1,3.7]$ and $[3.2,4.0]$.
The matched comparisons are also directionally stable: BAP-SQL minus SQL-RL
ranges from 2.5--2.7 points at XS and 2.2--2.7 at S, while BAP-SQL minus
BACM-RL ranges from 1.3--1.5 and 1.5--1.8 points.

\begin{table*}[t]
\centering
\small
\begin{tabular}{lrrrrr}
\toprule
Method & XS (\%) & S (\%) & M (\%) & L (\%) & \shortstack{Frontier\\Score (pp)} \\
\midrule
Raw agent & 17.9 & 19.1 & 19.7 & 19.9 & 19.2 \\
Dashboard & 17.7 & 17.6 & 17.4 & 19.0 & 17.8 \\
Adaptive post-hoc & 26.0 & 29.2 & 31.5 & 33.5 & 30.2 \\
\method{} SFT & 31.4 & 34.0 & 36.2 & 37.5 & 34.9 \\
SQL-RL (matched) & 32.2 & 35.2 & 38.7 & \textbf{40.4} & 36.7 \\
Context-only RL & 32.3 & 35.0 & 37.0 & 38.0 & 35.7 \\
BACM-RL (adapted) & 33.4 & 36.0 & 38.2 & 39.4 & 36.9 \\
\textbf{\method{} RL} & \textbf{34.8} & \textbf{37.6} & \textbf{39.2} & 39.0 & \textbf{37.9} \\
\bottomrule
\end{tabular}
\caption{Budgeted Success on the BIRD-derived dev set, pooled over three
adapters for trained methods. Frontier Score is the mean of the
trapezoidal interpolant over equally spaced ordinal XS--L positions; it is not
a probability or inferential endpoint. Adapter-seed Frontier Scores for
\method{} RL are 38.6/37.9/37.2 (sample standard deviation 0.70). At L,
matched SQL-RL has the
highest observed mean.}
\label{tab:frontier}
\end{table*}

\paragraph{RQ2: Quality and interaction cost.}
Both primary budgets pass the ordered quality-then-cost test
(Table~\ref{tab:endpoint}). Total tokens fall by 4.5\% at XS and 5.0\% at S,
or about 66 and 84 tokens per episode. Quality improves on 10 of 11 databases
at XS and 9 of 11 at S (one-sided sign-test $p\approx0.006$ and $0.033$);
token reductions hold on 10 of 11 databases at both budgets. Database-$t$
intervals, $[2.0,4.8]$ and $[2.1,5.1]$, preserve the direction. At L, the
token interval crosses zero.

Relative to BACM-RL, \method{} has slightly lower total-token use
($-1.3\%$ at XS and $-1.4\%$ at S), while BACM-RL keeps active context
4.1\% and 4.4\% smaller. The methods therefore occupy different
quality--cost points rather than dominating every resource.

Savings arise from shorter trajectories, not uniformly smaller outputs.
\method{} admits 13.8 and 30.2 result tokens at XS/S, slightly more than SFT,
but completes in 2.82 and 3.58 turns versus 2.93 and 3.84. On the
outcome-selected rescued subset, offline replay of discarded candidates reduces
median result tokens from 121 to 46 at XS and from 336 to 139 at S; VM steps
fall from 336K to 181K and from 804K to 451K. These subset diagnostics explain
how rewrites rescue binding cases, but pooled database work still increases by
about 0.8\%; we claim no database-compute saving.

\begin{table}[t]
\centering
\footnotesize
\begin{tabular}{@{}lrrr@{}}
\toprule
Budget & \shortstack{Quality\\diff.} & \shortstack{NI lower\\bound} &
\shortstack{Total-token change\\(95\% CI)} \\
\midrule
XS & +3.4 pp & +2.0 pp & -4.5\% [-5.9,-3.2] \\
S  & +3.6 pp & +2.1 pp & -5.0\% [-6.4,-3.7] \\
M  & +3.0 pp & +1.5 pp & -3.0\% [-4.2,-1.8] \\
L  & +1.5 pp & +0.2 pp & +0.2\% [-0.9,1.3] \\
\bottomrule
\end{tabular}
\caption{Paired RL--SFT endpoint. The non-inferiority (NI) criterion is met
when the one-sided 95\% database-bootstrap lower bound exceeds $-5$ pp. Both
XS and S must then satisfy the total-token-reduction criterion (paired 95\%
intervals entirely below zero). Intervals
resample 11 databases and condition on the three evaluated adapters. Quality
sign counts are 10/11 (XS) and 9/11 (S); token-reduction sign counts are 10/11
at both budgets. M/L are secondary.}
\label{tab:endpoint}
\end{table}

\paragraph{RQ3: Pressure and mechanism.}
Mixed single-channel budget vectors are sensitivity probes because the policy
was trained on the four joint levels. Around S, the \method{}--SFT gaps remain
positive when tightening only context, queries, result tokens, or database
work; the intervals overlap, so we do not rank channels. Result-token and
database-work pressure have the largest point estimates.

\paragraph{Training-level planning effect.}
The left panel of Table~\ref{tab:mechablate} is an independently locked rerun;
its cells should not be mixed with the main-frontier cells. In the locked
suite, SFT reaches 31.4/34.3, matched SQL-RL 32.2/35.1, no-planning RL
32.6/35.2, and full \method{} 34.5/37.2; the additional no-planning SQL-RL
cell reaches 32.1/34.5. Full-minus-no-planning is therefore 1.9/2.0 percentage
points, with paired 95\% intervals
$[0.7,3.1]/[0.8,3.2]$. Across seeds, the effect ranges from 1.7--2.1 points at
XS and 1.8--2.2 at S; every seed is positive. The locked SFT and SQL-RL cells
are within 0.3 and 0.1 points of their main-frontier counterparts, while the
full rerun is 0.3/0.4 points lower. Both action-matched SFT initializations
round to the displayed 31.4/34.3 before RL. The result identifies the combined
policy-visible estimate-and-rewrite package under a shared reward and
curriculum; it does not separate estimation from rewriting.

\paragraph{Reward--planning interaction.}
The completed factorial separates planning and reward within the locked suite.
Without planning, full reward improves over pure correctness by 0.5/0.7
points; with planning, it improves by 2.3/2.1 points. Conversely, planning adds
0.1/0.6 points under pure correctness and 1.9/2.0 under full reward. The
planning-by-reward interaction is therefore $+1.8/+1.4$ percentage points at
XS/S, indicating that efficiency shaping and the pre-execution interface act
as complements in this training setting.

\paragraph{Fixed-policy diagnostics.}
The right panel uses the separate main-run reference (34.8/37.6).
No-rewrite is worse than removing both actions, consistent with a frozen policy
receiving estimates it cannot act on. The shuffled placebo reaches 31.9/34.4;
its differences from no-interface are $+0.2/+0.2$, with intervals
$[-0.7,1.1]/[-0.8,1.2]$. These diagnostics show inference-time sensitivity
but can include distribution mismatch. Retrained no-efficiency shaping loses
1.9/2.0 points and uses 5.0\% more tokens; no-abstention changes success by
only 0.3 points.

\paragraph{Estimator behavior.}
Estimator p95 coverage is 98.8\% for rows, 97.0\% for result tokens, and 96.0\%
for database work. At least one policy-visible warning occurs in 54\% of XS
episodes and 48\% of S episodes. Of these, 71\%/66\% trigger rewrites, and
64\%/61\% of warned-and-rewritten episodes succeed; shield-only rejected
episodes recover in 55\%/52\%. These are descriptive conditional frequencies.
The shield records zero hard breaches by construction.

\begin{table*}[t]
\centering
\footnotesize
\begin{minipage}[t]{0.43\textwidth}
\centering
\textbf{A. Locked retraining suite}\par\smallskip
\begin{tabular}{@{}lrr@{}}
\toprule
Matched training condition & XS & S \\
\midrule
SFT & 31.4$\pm$0.3 & 34.3$\pm$0.3 \\
No-planning SQL-RL & 32.1$\pm$0.5 & 34.5$\pm$0.7 \\
Matched SQL-RL & 32.2$\pm$0.3 & 35.1$\pm$0.3 \\
No-planning RL & 32.6$\pm$0.2 & 35.2$\pm$0.2 \\
\textbf{Full \method{} RL (rerun)} & \textbf{34.5$\pm$0.4} & \textbf{37.2$\pm$0.4} \\
\bottomrule
\end{tabular}
\end{minipage}
\hfill
\begin{minipage}[t]{0.54\textwidth}
\centering
\textbf{B. Main-run diagnostics}\par\smallskip
\begin{tabular}{@{}lrrl@{}}
\toprule
Diagnostic & XS & S & $\Delta$ vs full (pp) \\
\midrule
Main-frontier reference & 34.8 & 37.6 & --- \\
No interface & 31.7 & 34.2 & $-3.1/-3.4$ \\
Shuffled estimate & 31.9 & 34.4 & $-2.9/-3.2$ \\
No rewrite & 30.9 & 33.0 & $-3.9/-4.6$ \\
Deterministic rewrite & 31.8 & 34.3 & $-3.0/-3.3$ \\
No efficiency shaping & 32.9 & 35.6 & $-1.9/-2.0$ \\
No abstention & 34.5 & 37.3 & $-0.3/-0.3$ \\
\bottomrule
\end{tabular}
\end{minipage}
\caption{Mechanism evidence at XS/S. Left: an independently locked matched
retraining suite, reported as mean$\pm$seed SD. Full \method{} exceeds
no-planning RL by 1.9/2.0 percentage points, with paired intervals
$[0.7,3.1]/[0.8,3.2]$; the planning-by-reward interaction is 1.8/1.4 points.
Right: diagnostics from the separate main run; all are fixed-policy except
no-efficiency shaping. Effects are computed only within their panel.}
\label{tab:mechablate}
\end{table*}

\begin{figure*}[t]
\centering
\includegraphics[width=0.98\textwidth]{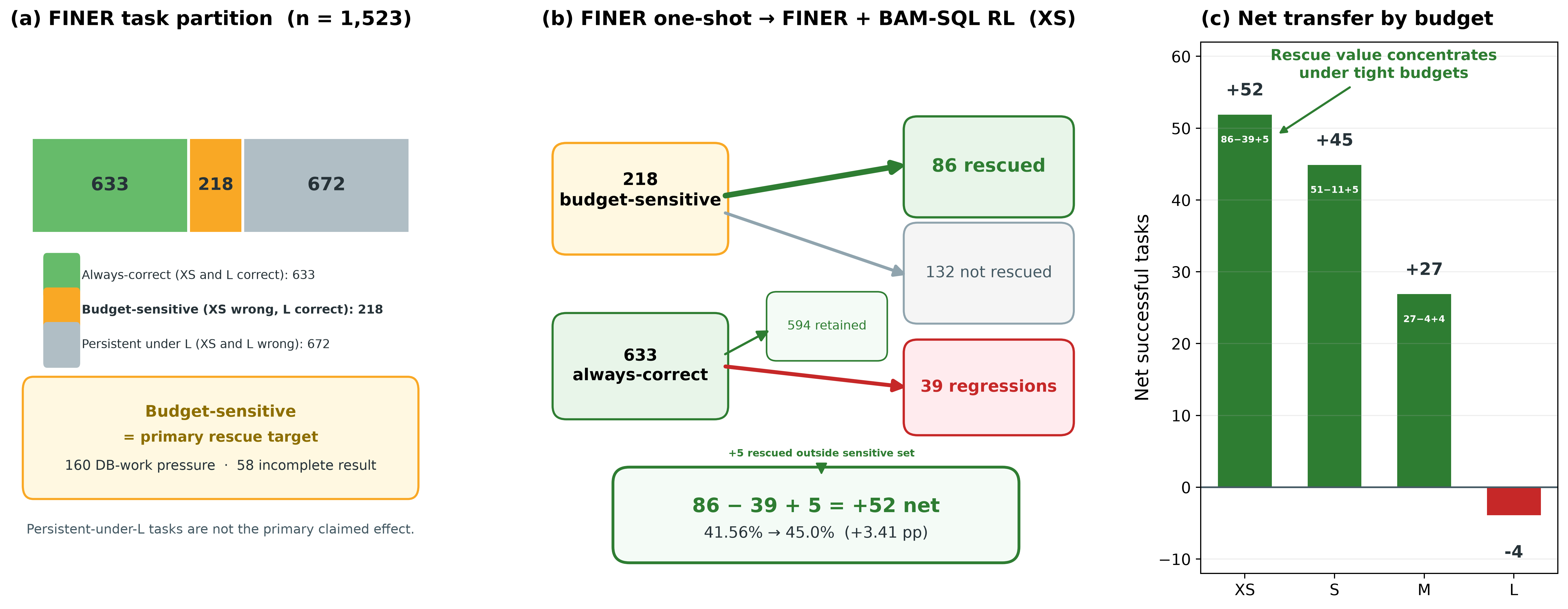}
\caption{Median-seed FINER-SQL task-level transfer; frontier point estimates
average three adapters. Across seeds, the XS net ranges from 46--58 tasks and
the S net from 40--51. The frozen one-shot query partitions
tasks into always-correct, budget-sensitive, and persistent-under-L groups
(left). At XS, \method{} rescues 86 budget-sensitive tasks, regresses 39
previously correct tasks, and solves 5 additional tasks, for a net gain of 52
(middle). Net gains concentrate at XS/S and reverse slightly at L (right).}
\label{fig:finer-rescue}
\end{figure*}

\paragraph{RQ4: Strong-backbone transfer.}
One-shot FINER already reaches
41.6\% and 48.3\% at XS/S. \method{} reaches 45.0\% and 51.3\%, with paired
one-shot differences of $[2.0,4.8]$ and $[1.6,4.4]$. At XS,
Figure~\ref{fig:finer-rescue} attributes the net gain to 86 rescued
budget-sensitive tasks, 39 regressions, and 5 additional successes. The net
gain shrinks at M and reverses slightly at L.
This pattern is seed-stable: at XS, 82--91 tasks are rescued, 39--40 regress,
and the net gain ranges from 46--58 tasks (3.0--3.8 points); at S, 47--56 are
rescued, 11--12 regress, and the net ranges from 40--51 tasks (2.6--3.3
points).
Planning raises total-token use by about 4\%: 1,530 (XS) and 1,550 (S) tokens
per episode versus 1,473 and 1,491 for one-shot FINER.
Relative to matched SQL-RL, the differences are
$+1.2$ $[0.3,2.1]$ and $+1.0$ $[0.1,1.9]$; relative to BACM-RL they are
$+1.0$ $[0.1,1.9]$ and $+0.7$ $[-0.2,1.6]$. The XS regressions primarily
follow unnecessary rewrites, motivating more conservative gating. Official EX
changes descriptively from 66.82\% to 66.7\%; we make no equivalence claim.

\paragraph{Boundary conditions and failures.}
The advantage is regime-specific. At L, correctness-oriented SQL-RL is stronger
on both backbones; pooled database work does not decrease. On a 7B coder model,
the XS/S gains attenuate to $+2.1$ $[0.9,3.3]$ and $+1.8$ $[0.6,3.0]$.
Thus the benefit transfers across model scale, but stronger SQL generation
leaves less headroom than the general 4B setting. On BIRD-INTERACT-LITE,
normalized
reward changes from 18.7\% to 19.8\% ($+1.1$ points $[-0.8,3.0]$) and tokens
change by $-3.5\%$ $[-7.0,0.0]$. Tightening or loosening the ladder by 25\%
yields XS/S mean gains of $+3.2$ $[0.8,5.6]$ and $+2.7$ $[0.4,5.0]$;
a common 10-turn ceiling retains $+3.2/+3.3$ points. Adding 15\% and 30\%
estimator noise changes success by $-0.4$ and $-1.8$ points, with zero hard
breaches. The shield guarantees feasibility, not optimality.

At XS, residual SQL-semantic and schema/value errors account for an estimated
37\% of tasks. Terminal rejection plus unjustified abstention accounts for
15\%, compared with 19.5\% for adaptive post-hoc control; remaining failures
come from insufficient evidence, tool selection, or output format. Labels use
a balanced, double-coded sample of 600 trajectories ($\kappa=0.78$).
\method{} therefore reduces budget-allocation errors but does not solve base
SQL competence.

\section{Limitations, Ethics, and Disclosure}

\paragraph{Scope.}
The primary study covers one general 4B backbone, one specialized SQL backbone,
11 dev databases, and three adapter seeds. Database-clustered intervals
condition on those adapters rather than marginalizing all training randomness.
The 7B check retains a smaller but positive effect, supporting transfer across
model scale. The BIRD-INTERACT-LITE interval still crosses zero, so transfer
to a different interactive task remains unestablished.

\paragraph{Resource trade-offs.}
The efficiency profile is backbone-dependent. On the general 4B model,
\method{} saves about 66--84 tokens per episode at XS/S. On FINER-SQL, the same
method spends about 4\% more total tokens for a quality gain, because the
one-shot baseline already executes a single SQL. \method{} also increases the
number of planning actions and does not reduce database work. Mixed
single-channel vectors are out-of-distribution sensitivity probes rather than
clean channel-specific interventions.

\paragraph{Comparison and mechanism limits.}
ContextBudget requires an environment adaptation because its original
action space manages search history rather than SQL observations. We keep its
released policy and learning procedure intact and freeze the SQL generator, but
a single harness cannot eliminate every implementation difference. The
comparison is strongest at the shared abstraction of budget-conditioned control
over incoming information.

The no-rewrite and no-estimate/rewrite rows are fixed-policy interventions and
can include off-policy mismatch. The cost-matched placebo more cleanly tests
inference-time dependence on estimator information. The retrained no-planning
control establishes a training-level contribution for the combined
policy-visible estimate-and-rewrite package. It does not isolate estimator
calibration or rewriting separately; a retrained shuffled-estimate placebo
would address part of this gap. The completed reward--planning factorial
identifies their interaction at the package level but remains specific to the
locked curriculum. Residual failures remain dominated by SQL semantics and
schema linking.

\paragraph{Ethics and disclosure.}
The system executes read-only SQL on local databases. Deployment still requires
least privilege, sensitive-column controls, and query auditing. Estimates can
encode dataset-specific cardinality patterns; a deployed catalog should be
rebuilt when schemas or data distributions change. Generative AI assisted
planning, drafting, and review; human authors remain responsible for the paper
and its claims.

\section{Conclusion}

Tool-using agents shape future observations through action choice. \method{}
instantiates this principle for agentic text-to-SQL by exposing pre-execution
cost information, learning when to issue a different evidence query, and
separating policy decisions from hard enforcement. Across general,
SQL-specialized, and larger generators, it improves tight-budget success; the
gain concentrates on budget-sensitive failures, and matched retraining provides
training-level evidence for policy-visible planning and its positive
interaction with efficiency shaping. The effect attenuates
as generator capability and available resources increase, reverses at the
loosest evaluated budget, and does not reduce database work. Thus
pre-execution observation planning is a targeted complement to stronger task
policies and post-hoc context control, and a broader design principle for
agents whose actions determine the cost and content of incoming observations.

\bibliographystyle{plainnat}
\bibliography{references}
\end{document}